\documentclass[sigconf, nonacm, screen=true, printacmref=false, printccs=false, printfolios=false]{acmart}
\renewcommand\footnotetextcopyrightpermission[1]{} % removes footnote with conference information in first column

\setcopyright{none}
\usepackage{xcolor}
\definecolor{thedarkblue}{RGB}{0,0,120} %104} % 180
\definecolor{mydarkblue}{rgb}{0,0.08,0.45} %ICML dark blue
\definecolor{darkblue}{rgb}{0,0.08,180}
\colorlet{TufteRed}{red!80!black}
\definecolor{theblue}{RGB}{0,0,180}
\colorlet{thered}{TufteRed}
      
\usepackage{microtype}
\usepackage{balance}

\usepackage{booktabs}
\usepackage{tabularx}

\usepackage{amsmath,amsthm}

\newcommand{\eat}[1]{\ignorespaces}
\usepackage{comment}

\usepackage{tikz}
\usepackage{verbatim}
\usetikzlibrary{arrows}
\usetikzlibrary{shapes,snakes}
\usetikzlibrary{decorations.pathmorphing} % noisy shapes
\usetikzlibrary{fit}					% fitting shapes to coordinates
\usetikzlibrary{backgrounds}	% drawing the background after the foreground

\usepackage{ragged2e}
\usepackage{multirow}
\usepackage{microtype}
\usepackage{balance}
\usepackage{setspace}

\graphicspath{{./}{./graphics/}}
\newcolumntype{H}{>{\setbox0=\hbox\bgroup}c<{\egroup}@{}}

\newcolumntype{R}[1]{>{\RaggedLeft\arraybackslash}} %p{#1}}
\newcolumntype{L}[1]{>{\RaggedRight\arraybackslash}} %p{#1}}

\newcommand{\eg}{\emph{e.g.}}
\newcommand{\ie}{\emph{i.e.}}

\AtBeginEnvironment{pmatrix}{\setlength{\arraycolsep}{2pt}}

\providecommand{\mat}[1]{\boldsymbol{\mathrm{#1}}}%
\renewcommand{\vec}[1]{\boldsymbol{\mathrm{#1}}}

\DeclareMathOperator{\hugeE}{\mbox{\huge\raise-0.3ex\hbox{E}}}
\DeclareMathOperator{\p}{\mathbb{P}}
\DeclareMathOperator{\hugep}{\mbox{\huge\raise-0.3ex\hbox{$\p$}}}

\providecommand{\mA}{\ensuremath{\mat{A}}}

\providecommand{\mW}{\ensuremath{\mat{W}}}

\providecommand{\vc}{\ensuremath{\vec{c}}}

\providecommand{\ve}{\ensuremath{\vec{e}}}

\providecommand{\vx}{\ensuremath{\vec{x}}}

\usepackage{graphbox}
\usepackage{svg}
\usepackage{nicefrac}
\usepackage[noend]{algpseudocode}

\usepackage{amsfonts}
\usepackage[linesnumbered, ruled]{algorithm2e}
\usepackage{xcolor}

\usepackage{multirow}
\usepackage{caption}
\usepackage{subcaption}
\usepackage[normalem]{ulem}
\DeclareMathAlphabet{\mathbcal}{OMS}{cmsy}{b}{n}
\usepackage{amsmath}
\usepackage{mathrsfs}
\usepackage{comment}

\usepackage{bm}
\usepackage{bbm}

\usepackage{enumitem}
\AtBeginDocument{
  \providecommand\BibTeX{{
    \normalfont B\kern-0.5em{\scshape i\kern-0.25em b}\kern-0.8em\TeX}}}

\copyrightyear{2023}
\acmYear{2023}
\setcopyright{none}
\acmPrice{}
\acmISBN{}
\acmConference[ArXiv '23]{}{}{}
\acmBooktitle{}
\acmPrice{}
\acmDOI{}

\begin{document}

\title{PersonaSAGE: A Multi-Persona Graph Neural Network}

\settopmatter{authorsperrow=3}

\author{Gautam Choudhary}
\affiliation{
  \institution{Adobe Research}
}
\email{gc.iitr@gmail.com}

\author{Iftikhar Ahamath Burhanuddin}
\affiliation{
  \institution{Adobe Research}
}
\email{burhanud@adobe.com}

\author{Eunyee Koh}
\affiliation{
  \institution{Adobe Research}
}
\email{eunyee@adobe.com}

\author{Fan Du}
\affiliation{
  \institution{Adobe Research}
}
\email{fdu@adobe.com}

\author{Ryan A. Rossi}
\affiliation{
  \institution{Adobe Research}
}
\email{ryrossi@adobe.com}

\begin{abstract}
Graph Neural Networks (GNNs) have become increasingly important in recent years due to their state-of-the-art performance on many important downstream applications. Existing GNNs have mostly focused on learning a single node representation, despite that a node often exhibits polysemous behavior in different contexts. In this work, we develop a persona-based graph neural network framework called PersonaSAGE that learns multiple persona-based embeddings for each node in the graph. Such disentangled representations are more interpretable and useful than a single embedding. Furthermore, PersonaSAGE learns the appropriate set of persona embeddings for each node in the graph, and every node can have a different number of assigned persona embeddings. The framework is flexible enough and the general design helps in the wide applicability of the learned embeddings to suit the domain. We utilize publicly available benchmark datasets to evaluate our approach and against a variety of baselines. The experiments demonstrate the effectiveness of PersonaSAGE for a variety of important tasks including link prediction where we achieve an average gain of 15\% while remaining competitive for node classification. Finally, we also demonstrate the utility of PersonaSAGE with a case study for personalized recommendation of different entity types in a data management platform.
\end{abstract}




\maketitle

\begin{figure}[t!]
\vspace{2mm}
\centering
\includegraphics[width=1.0\linewidth]{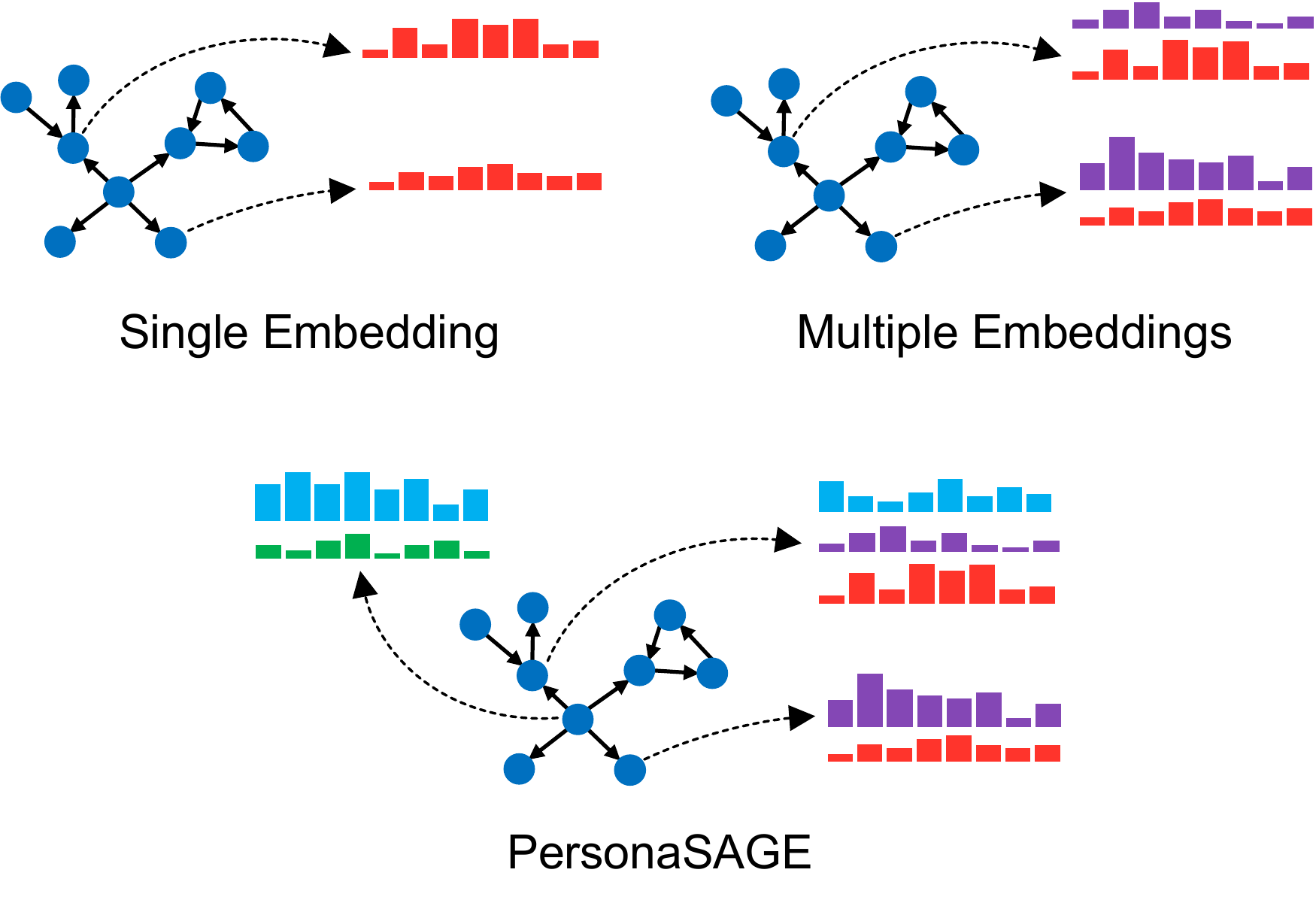}
\vspace{-6mm}
\caption{
Overview of the learning objectives of PersonaSAGE and prior work.
While previous work has mainly focused on learning single embedding per node, a few recent works learn multiple embeddings per node.
In contrast, PersonaSAGE not only learns multiple persona embeddings but also learns the appropriate number of persona embeddings per node.
}
\label{fig:novelty-learning-objectives}
\vspace{-3mm}
\end{figure}

\section{Introduction}
In recent years, Graph Neural Networks (GNNs) have become increasingly important due to their state-of-the-art performance on many important downstream applications including link prediction and node classification~\cite{wu-gnn-survey,zhou2020graph,rossi2020proximity,velivckovic2017graph,deepGL,damke-gnn, gnn-wl-maron}.
Existing GNNs have mostly focused on learning a single node embedding (or representation)~\cite{zhou2020graph,rossi2020proximity}, despite that a node often exhibits polysemous behavior in different contexts~\cite{ahmed2017edge}.
For instance, an individual may have many different personas, e.g., a user may be a researcher, father, coach, and activist~\cite{rossi2014role,liu2019single}. 
These personas may be fundamentally different or even impossible for other individuals.
Each one of these general personas may also have sub-personas that capture specific behaviors and characteristics of each, which we call sub-personas.
However, existing methods are unable to learn such sets of persona embeddings for each node in the graph.

To address this limitation, we develop a persona-based graph neural network framework called PersonaSAGE that learns a set of embeddings for each node in the graph.
Furthermore, PersonaSAGE learns the appropriate set of persona embeddings for each node in the graph, and every node can have a different number of assigned personas.
In Figure~\ref{fig:novelty-learning-objectives}, we provide an intuitive overview comparing the learning objective of PersonaSAGE to previous work. 
Consider the real-world scenario of roles in a data management platform. The central node is that of a data science manager and is represented by two embeddings: blue and green. The green embedding denotes the role of the manager (and their sub-roles). The other nodes being individual contributors such as a data analyst, data engineer or data scientist are represented by embeddings other than the green embedding based on their respective roles. The manager doubles as one of these roles, for instance, data scientist, and is additionally represented by a blue embedding. This scenario illustrates the need for different nodes in the graph to be represented by a varying number of embeddings depending on their role and community membership. Each set of embeddings collectively represents the node. 
Notably, as shown in Figure~\ref{fig:novelty-learning-objectives}, every node can have a different set of embeddings along with a different number of assigned embeddings as well. 
For instance, one node in Figure~\ref{fig:novelty-learning-objectives} has 3 embedding vectors compared to the other nodes that have only 2 embeddings. Furthermore, the nodes with 2 embedding vectors are also distinct, representing different personas (behaviors and structural characteristics) altogether, which is shown by the color of the embedding vectors in Figure~\ref{fig:novelty-learning-objectives}.

The experiments demonstrate the effectiveness of our approach for link prediction and node classification.
Overall, we find that PersonaSAGE always outperforms the other methods across all graphs and prediction tasks.
For link prediction, PersonaSAGE achieves a mean gain of 15\% over the best baseline method across all graphs.
PersonaSAGE also performs well for node classification achieving a mean gain of 1\% and up to 17\% over the other methods.
Furthermore, we also conducted a case study in Section~\ref{sec:case-study} where PersonaSAGE is applied for personalized recommendation of queries, attributes, and datasets. 
In this case study, PersonaSAGE learns multiple embeddings per node from a large heterogeneous graph derived from the usage logs of users (from a data management platform).
Notably, PersonaSAGE significantly outperforms the other methods, achieving a gain of $19.2\%$, $15.7\%$, and $21.7\%$ in AUC over the best baseline method for query, attribute, and dataset recommendation tasks, respectively.
These results demonstrate the effectiveness of PersonaSAGE and its ability to learn sets of embeddings for every node in the graph that appropriately capture the contextual behavior and personas of the nodes.

The main contributions are as follows:
\begin{itemize}

\item \textbf{Problem Formulation:} We introduce the problem of automatically learning sets of embeddings for each node in the graph, which may also be of different sizes depending on the structural characteristics surrounding the node.

\item \textbf{Novel Framework:} This work develops PersonaSAGE, a novel graph neural network framework that learns multiple embeddings for every node, which are also flexible in size. Hence, PersonaSAGE may learn 2 embeddings for one node and 3 embeddings for another, thus, every node is automatically assigned the appropriate set of embeddings.

\item \textbf{Effectiveness:} Through comprehensive experiments, PersonaSAGE is shown to be extremely effective for a wide variety of application tasks including link prediction and node classification. Notably, PersonaSAGE outperforms the other methods across a wide variety of graphs.
Finally, we also demonstrate the utility of PersonaSAGE on a case study where we apply these techniques for personalized recommendation of queries, datasets, and attributes in a data management platform.

\end{itemize}

\section{\NoCaseChange{PersonaSAGE} Framework} \label{sec:approach}
In this section, we describe the proposed approach called PersonaSAGE. We first provide an overview and then detail the algorithm for the generation of multiple sets of persona embeddings. Later, we describe specific choices involved in our approach such as persona assignments and the kind of aggregations involved. An end-to-end optimization problem is designed to learn the parameters for downstream applications such as link prediction and node classification.

\subsection{Problem Formulation} \label{sec:problem-formulation}
In this work, we investigate the new problem formally described below. Given an arbitrary graph $G=(\mathbcal{V}, \mathbcal{E})$ along with its adjacency matrix $\mA$, the problem is to learn multiple embeddings $\mathbcal{X}_v=\{\vx_1, \vx_2, \ldots\}$ for every node $v \in \mathbcal{V}$ where (i) for any two different nodes $u,v \in \mathbcal{V}$, $|\mathbcal{X}_u| \not=|\mathbcal{X}_v|$ may hold and (ii) $|\mathbcal{X}_u \cup \mathbcal{X}_v| = |\mathbcal{X}_u|+|\mathbcal{X}_v|$. Intuitively, (i) implies that different nodes can have different number of embeddings whereas (ii) implies that the embeddings learned for any node in the graph are unique (non-identical), otherwise $\exists~ u,v \in \mathbcal{V}, |\mathbcal{X}_u \cup \mathbcal{X}_v| < |\mathbcal{X}_u|+|\mathbcal{X}_v|$ would hold. Let $K = \max_{v \in \mathbcal{V}} |\mathbcal{X}_v|$ denote the maximum number of embeddings per node in $G$ such that $K \ll n$, where $n$ is the number of nodes in $G$. Without loss of generality, all embeddings are assumed to be of fixed size $D$, \ie, $(\vx_i \in \mathbcal{X}_v) \in \mathbb{R}^D$.
Intuitively, each learned embedding $\vx_i \in \mathbcal{X}_v$ represents a set of ``sub-personas'' for a given ``structural context'', which are also automatically learned from the data.\footnote{This is in contrast to recent work~\cite{epasto2019single} that assumes the different contexts are given as input and simply learns an embedding for each one.}
As an aside, in this work, uppercase letters in calligraphic bold font denote sets,  $\mathbcal{X}$.

\subsection{Overview}
\begin{figure}[t!]
    \centering
    \vspace{-2mm}
    \includegraphics[width=1.0\linewidth]{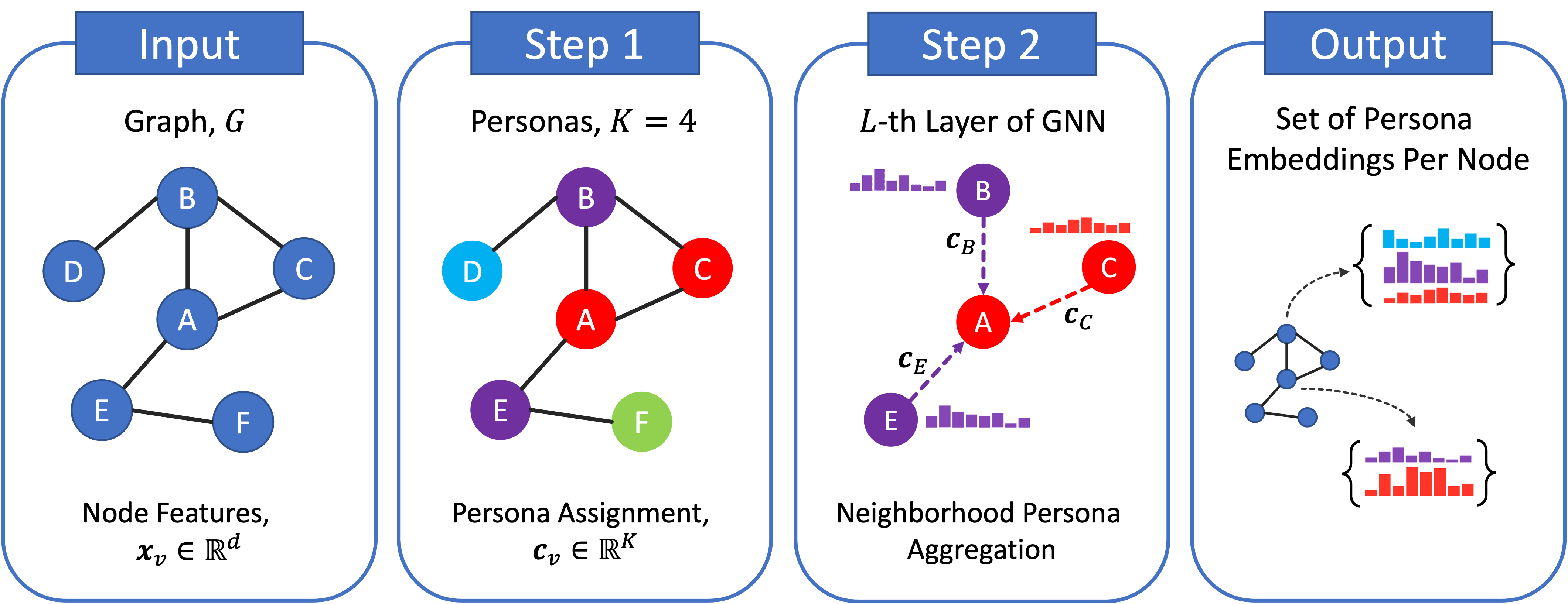}
    \vspace{-4mm}
    \caption{Overview of our proposed approach, PersonaSAGE. 
    }
    \label{fig:overview}
    \vspace{-2mm}
\end{figure}

Figure~\ref{fig:overview} highlights the given inputs and desired outputs of the system and an overview of our proposed model, PersonaSAGE. As shown, given a sample graph comprising $n=6$ nodes and their node embeddings, we compute a reference persona label for each node through a clustering algorithm by assuming $K$ sets of personas in the input graph data. Suppose that $K=4$ personas are considered and each persona is represented by a distinct color. To visualize, assume the output cluster labels as highlighted in \textit{Step~1} of the figure. We can convert them into one-hot characteristic encodings, $\vc_v \in \mathbb{R}^K$ for each node $v$ in the graph. We refer to these encodings as initial persona membership vectors whose each element defines the degree of association of every persona to the given node. For each node, by using an initial node embedding and persona membership vector as input, each layer of PersonaSAGE iteratively updates them based on the neighboring nodes as highlighted in \textit{Step~2} in the figure. The model finally yields a set of persona embeddings and a corresponding updated persona membership vector for each node. The desired output for this example is represented in the last block of the figure where node $A$ has a higher membership for the \textit{red} persona than the \textit{violet} and has $2$ corresponding persona embeddings as intuitive from its neighborhood. Similarly, node $B$ has a persona embedding for each of the $3$ persona memberships, namely \textit{red}, \textit{violet}, and \textit{blue} with proportions relative to its neighborhood. Our model is capable of finding the appropriate number of personas for each node. In the next section, we describe the update procedures for the persona membership vector as well as their corresponding persona embeddings for each node.

\subsection{Algorithm}
The core idea behind the PersonaSAGE algorithm is to represent each node as a set of persona embeddings where the cardinality of this set may differ for different nodes. Let $K$ denote the maximum number of persona embeddings (or embedding vectors) per node. For instance, $K=4$ in Figure~\ref{fig:novelty-learning-objectives} as a user may have four embedding vectors since there are four different colors denoting the various persona embedding vectors. Furthermore, let $D$ denote the maximum embedding size for the $K$ different persona embeddings that a node in $G$ can be assigned. Our approach models the persona representation of a node $v$ as a combination of two aspects: (i) a persona membership vector $\vc_{v} \in \mathbb{R}^K$ of unit length ($\|\vc_{v}\|_1 = 1$) denoting proportionate degree of membership for each persona, and (ii) the corresponding persona embeddings for non-zero persona memberships,
$\mathbcal{X}_{v} = \{\vx_i ~|~ \vc_{v, i} > 0, \vx_i \in \mathbb{R}^D \} $.
We represent $\vc_{v,i}$ to denote the degree of engagement or association of a particular node $v$ to a persona $i \in {1, \ldots, K}$ while $\vx_i$ denotes the corresponding persona embedding in a $D$ dimensional vector space.

The forward propagation algorithm for PersonaSAGE is formally defined in Algorithm~\ref{algo:personasage}.
Given a graph, $G (\mathbcal{V}, \mathbcal{E})$, its initial set of node embeddings $\{\textbf{x}_v, \forall v \in \mathbcal{V}\}$ and the number of personas $K$ to be computed, along with the given set of initial membership vectors $\{\textbf{c}_v ~|~ v \in \mathbcal{V} \}$ as prior knowledge, the algorithm finds multiple (and different) sets of persona embeddings $\mathbcal{X}_v$ for each node $v$ in the graph. We also provide a strategy to learn membership vectors in the next section, in case the prior knowledge is not known. 

The model initializes each persona embedding for a given node $v \in \mathbcal{V}$ from the same input node features of that node to avoid any bias in Line 1. For each layer $l$ of the network, the algorithm performs two updates. The first update happens for persona membership vector $\textbf{C}_v^l$ by aggregating information from its neighbors $\mathbcal{N}(v)$ and then applying normalization in Line 5 and 6 respectively. 
We assume that the persona membership vector of a node is likely to be influenced by that of its neighbors and we update it at each layer of the network. This update strategy is based on the intuition that the \textit{persona} of a node is a function of itself and that of its neighbors where the neighborhood of a node is defined as its directly connected nodes.
The next update happens separately for each persona, where the updated membership vector is used to condition the neighboring persona embeddings of the previous layer $\{\textbf{X}_{u,i}^{l-1}, u \in \mathbcal{N}(v)\}$ for a reference node $v$ before applying a suitable aggregator function as described in Line 10. Intuitively, for the $i$-th persona embedding of a node is updated based on $i$-th persona embeddings of its neighbors conditioned upon their persona memberships $\textbf{C}_{u,i}^{l}$. We discuss various alternatives for aggregation functions in a subsequent section. 
The aggregated information from neighbors' $i$-th persona $\textbf{h}_{\mathbcal{N}(v),i}^l$ is then
concatenated with the node's own $i$-th persona embedding $\{\textbf{X}_{v,i}^{l-1}\}$ and passed through a neural network (affine transformations followed by a non-linear activation function $\sigma(\cdot)$ such as Sigmoid) where $\mW^l$ denote learnable weight matrices of this neural network.
Note that a single neural network is trained per layer and is the same for each persona. Both the update steps could be individually parallelized at the node level for efficient computation. Finally, the set of embeddings for which persona memberships are non-zero is returned. To reiterate, we aim to disentangle the node embeddings as a set of multiple persona embeddings conditioned by the persona membership vector.

The time complexity of the forward propagation phase of Algorithm~\ref{algo:personasage} is $O(L \cdot n \cdot K \cdot d)$, where $n$ is the cardinality of $\mathbcal{V}$, and $d$ is an upper bound of the maximum degree of the nodes. This time complexity can be significantly reduced by accounting for matrix computation and parallelism. 
We examine various clustering algorithms (\eg, KMeans, Ward) along with a choice of aggregation functions in an ablation study in Section~\ref{sec:exp}.

\begin{algorithm}
    \caption{PersonaSAGE Forward Propagation}\label{algo:personasage}
    \SetKwInOut{KwInput}{Input}
    \SetKwInOut{KwOutput}{Output}
    \SetCommentSty{mycommentfont}
    
    % Input
    \KwInput{Graph $G(\mathbcal{V}, \mathbcal{E})$; input features $\{\textbf{x}_v, \forall v \in \mathbcal{V}\}$; number of clusters $K$; membership vectors $\{ \textbf{c}_{v} \in \mathbb{R}^K, \forall v \in \mathbcal{V} \}$; number of layers $L$}
    % Output
    \KwOutput{Persona Embeddings $\mathbcal{X}_v, \forall v \in \mathbcal{V}$}
    
    % Initialisation
    \tcc{Initialisation}
    $\textbf{X}_{v, i}^{0} = \textbf{x}_v, \forall i \in \{1, \ldots, K\}, \forall v \in \mathbcal{V}$ \;
    $\textbf{C}_v^{0} = \textbf{c}_v, \forall v \in \mathbcal{V}$ \;
    
    % for each layer
    \For{$l \gets 1$ \KwTo $L$}{
        \tcc{Persona Membership Update}
        % for each node
        \For{$v \in \mathbcal{V}$}{
            % Membership Vector Update
            $\textbf{C}_v^{l} = \textbf{C}_v^{l-1} + \sum_{u \in \mathbcal{N}(v)} \textbf{C}_u^{l-1}$ \;
            $\textbf{C}_v^{l} = \textbf{C}_v^{l} / \| \textbf{C}_v^{l} \|_1$ \;
        }
        % for each node
        \tcc{Persona Embeddings Update}
        \For{$v \in \mathbcal{V}$}{
            % Embedding Vector Update
            \For{$i \gets 1$ \KwTo $K$}{
                $ \textbf{h}_{\mathbcal{N}(v), i}^{l} {=} f_{aggregate}^{l}\left(\left\{ (\textbf{C}_{u, i}^{l}, \textbf{X}_{u, i}^{l-1}) | u \in \mathbcal{N}(v)\right\}\right)$ \;
                $ \textbf{X}_{v, i}^{l} = \sigma \left(\mW^{l} \cdot CONCAT\left[\textbf{X}_{v, i}^{l-1}, \textbf{h}_{\mathbcal{N}(v), i}^{l}\right] \right) $ \;
            }
        }
    }
    $\mathbcal{X}_v = \left\{ \textbf{X}_{v,i} ~|~ \textbf{C}_{v, i} > 0, \forall i \in \{1, \ldots, K\}\right\}$ \;
    \KwRet{$\mathbcal{X}_v$} \;
\end{algorithm}

\subsection{Persona Assignment}
We now describe how to obtain initial persona membership vectors which our algorithm relies on as prior knowledge. We pose it as a clustering problem by partitioning the nodes in a graph $G$ into $K$ mutually exclusive sets. More formally, a node $v$ belonging to cluster $i$ will have a persona membership vector $\textbf{c}_v = \ve_i \in \{0,1\}^K$ , \ie, a one-hot encoded vector. We investigate various clustering algorithms such as KMeans, Spectral Clustering, Ward's Hierarchical Clustering, etc. for partitioning nodes in a graph. The clustering outputs a label denoting a hard assignment of a persona to each node. Our algorithm leverages this static assignment and automatically learns new personas for every node based on their neighborhood.

\subsection{Aggregator Functions}
An aggregator function is used to collect and transmit important and combined information from neighboring nodes $\mathbcal{N}(v)$ to a given node $v$. Since this information aggregation is not tied to a specific order, the aggregator functions are to be order invariant. We investigate three kinds of aggregator functions that are used to combine the persona information from neighborhood nodes $\mathbcal{N}(v)$ to a reference node $v$. Each describes how the persona membership value $c \in \mathbb{R}_{\geq 0}$ combine with its corresponding persona embedding $\textbf{x}_{u} \in \mathbb{R}^D$ and interact in the neighborhood $\mathbcal{N}(v)$ of node $v$.

\subsubsection{Mean} This aggregation performs an average operation of the information collected from neighbors, i.e.,
\[
    f_{aggregate}\left(\left\{ (c, \textbf{x}_{u}) ~|~ u \in \mathbcal{N}(v)\right\}\right) = \frac{\sum_{u \in \mathbcal{N}(v)} c \cdot \textbf{x}_u}{|\mathbcal{N}(v)|}
\]

\subsubsection{Sum} This aggregation sums up the information collected from neighbors, i.e.,
\[
    f_{aggregate}\left(\left\{ (c, \textbf{x}_{u}) ~|~ u \in \mathbcal{N}(v)\right\}\right) = \sum_{u \in \mathbcal{N}(v)} c \cdot \textbf{x}_u
\]

\subsubsection{Max} This aggregation performs the element-wise max operation of the neighboring node vectors, i.e.,
\[
    f_{aggregate}\left(\left\{ (c, \textbf{x}_{u}) ~|~ u \in \mathbcal{N}(v)\right\}\right) = [\max_{u \in \mathbcal{N}(v)}(c \cdot \textbf{x}_{u,1}), ...]
\]

\subsection{Training and Optimization}
Until now, we have discussed sets of persona embeddings with different cardinalities. For sake of practical usage in downstream tasks, we may need to obtain a single embedding of fixed length for each node. We can use multiple aggregation strategies. A straightforward option is an orderly concatenation of all $K$ persona embeddings:
\[ 
    \Tilde{\textbf{X}}_v = \textbf{X}_{v,1} \oplus \textbf{X}_{v,2} \oplus \ldots \oplus \textbf{X}_{v,K} 
\]
where we do not condition them with persona memberships. Another way is to explicitly condition them by scaling the embeddings with persona membership vector before concatenation:
\[ 
    \Tilde{\textbf{X}}_v = (\textbf{C}_{v,1}\cdot\textbf{X}_{v,1}) \oplus (\textbf{C}_{v,2}\cdot\textbf{X}_{v,2}) \oplus \ldots \oplus (\textbf{C}_{v,K}\cdot\textbf{X}_{v,K}) 
\]
and return this embedding than the set of persona embeddings which is shown in Line 15 in Algorithm~\ref{algo:personasage}. We adopt, the above formulation for computing final node embeddings in the experiments. Here, $\oplus$ denotes the concatenation operation.

Now, this vector can be easily used for downstream tasks. We describe two benchmark tasks: link prediction and node classification. For link prediction, the similarity score between two nodes $u, v$ is defined as the inner vector product: $sim(u, v) = \Tilde{\textbf{X}}_u \odot \Tilde{\textbf{X}}_v$.
For node classification, the node embeddings $\Tilde{\textbf{X}}_v$ can be fed to a neural network to learn node labels. In both cases, a standard loss function, such as cross-entropy loss, can be readily used in backpropagation to learn the weight parameters of the PersonaSAGE model.

\begin{table}[h!]
    \centering
    \caption{Dataset Statistics. 
    }
    \label{tab:global-data-stats}
    \vspace{-2mm}
    \begin{tabular}{lcccc}
    \toprule
    \textbf{Dataset} &  \textbf{\#Classes} &  \textbf{\#Nodes} &  \textbf{\#Edges}  & \textbf{\#Features}\\
    \midrule
    Citeseer    & 6     & 3,327     & 4,732     & 3703 \\
    Cora        & 7     & 2,708     & 5,429     & 1433 \\
    PubMed      & 3     & 19,717    & 44,338    & 500 \\
    \bottomrule
    \end{tabular}
\end{table}

\begin{table*}[h!]
    \centering
    \vspace{-2mm}
    \caption{
    Link Prediction Results.
    }
    \label{tab:link-prediction-main}
    \vspace{-4mm}
    \begin{tabular}{l cc cc cc}
    \toprule
    \multicolumn{1}{l}{\textbf{}}   & \multicolumn{2}{c}{\textbf{Citeseer}} & \multicolumn{2}{c}{\textbf{Cora}} & \multicolumn{2}{c}{\textbf{PubMed}} \\
    \cmidrule(lr){2-3}
    \cmidrule(lr){4-5}
    \cmidrule(lr){6-7}
    \textit{\textbf{Model}}         & \textbf{Raw Feat}       & \textbf{Random Feat}    & \textbf{Raw Feat}     & \textbf{Random Feat}  & \textbf{Raw Feat}      & \textbf{Random Feat}   \\
    \midrule
    \textit{Cheb}                   & 0.733 $\pm$ 0.03             & 0.563 $\pm$ 0.09             & 0.686 $\pm$ 0.06           & 0.603 $\pm$ 0.10            & 0.803 $\pm$ 0.03            & 0.778 $\pm$ 0.01            \\
    \textit{Edge}                   & 0.639 $\pm$ 0.03             & 0.576 $\pm$ 0.02             & 0.515 $\pm$ 0.02           & 0.475 $\pm$ 0.04           & 0.80 $\pm$ 0.00               & 0.732 $\pm$ 0.01            \\
    \textit{GAT}                    & 0.734 $\pm$ 0.02             & 0.661 $\pm$ 0.02             & 0.678 $\pm$ 0.01           & 0.661 $\pm$ 0.01           & 0.743 $\pm$ 0.01            & 0.699 $\pm$ 0.01            \\
    \textit{GCN}                    & 0.711 $\pm$ 0.01             & 0.682 $\pm$ 0.01             & 0.682 $\pm$ 0.01           & 0.676 $\pm$ 0.01           & 0.791 $\pm$ 0.00             & 0.726 $\pm$ 0.01            \\
    \textit{GraphSAGE}         & 0.816 $\pm$ 0.01             & 0.802 $\pm$ 0.01             & 0.667 $\pm$ 0.02           & 0.618 $\pm$ 0.02           & 0.747 $\pm$ 0.01            & 0.713 $\pm$ 0.01            \\
    \textit{SG}                     & 0.705 $\pm$ 0.01             & 0.671 $\pm$ 0.01             & 0.681 $\pm$ 0.01           & 0.674 $\pm$ 0.02           & 0.788 $\pm$ 0.01            & 0.723 $\pm$ 0.00             \\
    \textit{TAG}                    & 0.822 $\pm$ 0.02             & 0.798 $\pm$ 0.02             & 0.718 $\pm$ 0.00            & 0.717 $\pm$ 0.00            & 0.829 $\pm$ 0.00             & 0.774 $\pm$ 0.01            \\
    \midrule
    \textit{PersonaSAGE}  
    & \textbf{0.872 $\pm$ 0.02}             &  \textbf{0.907 $\pm$ 0.00}              & \textbf{0.828 $\pm$ 0.01}           & \textbf{0.911 $\pm$ 0.01}  & \textbf{0.950 $\pm$ 0.00}              & \textbf{0.871 $\pm$ 0.02}  \\
    \bottomrule
    \end{tabular}
    \vspace{-0mm}
\end{table*}

\section{Experiments}\label{sec:exp}
In this section, we design experiments to investigate the effectiveness of PersonaSAGE for link prediction (Section~\ref{sec:link-prediction}) and node classification (Section~\ref{sec:exp-node-classification}).
We then conduct an ablation study in Section~\ref{sec:ablation-study}.

\subsection{Experimental setup} \label{sec:exp-setup}

\subsubsection{Data} 
We consider the following citation network datasets for experiments: Cora, Citeseer, PubMed~\cite{sen2008collective}. These are datasets are a network of scientific publications (nodes), connected via a citation relationship, \ie, one publication cites another. Each node in the datasets is represented by a fixed-length vector derived using a bag-of-words representation of the document. The count of nodes, edges, and features is shown in Table~\ref{tab:global-data-stats}. The edge weights, if any, are not considered and we assume the graphs as undirected for sake of simplicity. The usage of these datasets is for two downstream tasks explained in the subsequent sections. Given two publications (an edge), the model could learn to predict if one cites another (link prediction) or given a publication (node), the model could learn to predict their classes as described in Appendix~\ref{sec:exp-setup-details}. These features are normalized (min-max) to obtain a range between 0 and 1. For our approach, we initialize the persona membership vector as the one-hot encoding of cluster labels obtained by applying a clustering algorithm on the normalized features. Since the clustering algorithm usually requires the number of clusters ($K$) to be supplied as input, we assume the number of personas is equivalent to the number of classes found in the data for sake of simplicity. This could be set to any other value as well.

\begin{table*}[h!]
    \centering
    \vspace{-0mm}
    \caption{Ablation study link prediction results investigating different variants of our PersonaSAGE framework.
    }
    \label{tab:link-prediction-ablation-study}
    \vspace{-4mm}
    \begin{tabular}{l cc cc cc}
    \toprule
    \multicolumn{1}{l}{\textbf{}}   & \multicolumn{2}{c}{\textbf{Citeseer}} & \multicolumn{2}{c}{\textbf{Cora}} & \multicolumn{2}{c}{\textbf{PubMed}} \\
    \cmidrule(lr){2-3}
    \cmidrule(lr){4-5}
    \cmidrule(lr){6-7}
    \textit{\textbf{Model}}         & \textbf{Raw Feat}       & \textbf{Random Feat}    & \textbf{Raw Feat}     & \textbf{Random Feat}  & \textbf{Raw Feat}      & \textbf{Random Feat}   \\
    \midrule
    \textit{PersonaSAGE-KM ($K=1$)}       & 0.753 $\pm$ 0.00              & 0.747 $\pm$ 0.00              & 0.740 $\pm$ 0.02            & 0.709 $\pm$ 0.01           & 0.939 $\pm$ 0.00             & 0.843 $\pm$ 0.01            \\
    \textit{PersonaSAGE-KM-max}             & 0.823 $\pm$ 0.05             & 0.863 $\pm$ 0.00              & 0.748 $\pm$ 0.04           & 0.869 $\pm$ 0.00            & 0.880 $\pm$ 0.01             & 0.822 $\pm$ 0.00             \\
    \textit{PersonaSAGE-KM-sum}             & 0.789 $\pm$ 0.05             & 0.885 $\pm$ 0.01             & 0.763 $\pm$ 0.03           & 0.884 $\pm$ 0.01           & 0.881 $\pm$ 0.01            & 0.834 $\pm$ 0.01            \\
    \textit{PersonaSAGE-KM}            & \textbf{0.881 $\pm$ 0.02}    & \textbf{0.916 $\pm$ 0.01}    & \textbf{0.857 $\pm$ 0.01}  & 0.908 $\pm$ 0.01           & \textbf{0.951 $\pm$ 0.00}    & \textbf{0.885 $\pm$ 0.00}             \\
    \textit{PersonaSAGE-Birch} & 0.872 $\pm$ 0.02             & 0.908 $\pm$ 0.00              & 0.833 $\pm$ 0.01           & 0.909 $\pm$ 0.01           & \textbf{0.951 $\pm$ 0.00}    & \text{0.871 $\pm$ 0.02}   \\
    \textit{PersonaSAGE-Spec}  
    & 0.867 $\pm$ 0.02             & 0.902 $\pm$ 0.02             & 0.828 $\pm$ 0.01           & 0.878 $\pm$ 0.03           & -                      & 0.819 $\pm$ 0.01            \\
    \textit{PersonaSAGE}  
    & 0.872 $\pm$ 0.02             & 0.907 $\pm$ 0.00              & 0.828 $\pm$ 0.01           & \textbf{0.911 $\pm$ 0.01}  & 0.950 $\pm$ 0.00              & \text{0.871 $\pm$ 0.02}  \\
    \bottomrule
    \end{tabular}
\end{table*}

\subsubsection{Baselines}
We use the following methods for comparison:
ChebNet (Cheb)~\cite{defferrard2016convolutional}, Graph Convolution Network (GCN)~\cite{kipf2016semi}, GraphSAGE~\cite{hamilton2017inductive}, Graph Attention Networks (GAT)~\cite{velivckovic2017graph}, Topology Adaptive Graph convolutional networks (TAG)~\cite{du2017topology}, EdgeConv (Edge)~\cite{wang2019dynamic}, Simple Graph convolution (SGN)~\cite{wu2019simplifying}. See Appendinx~\ref{sec:exp-setup-details} for further details.
We use a standard implementation of the baselines, as provided by Deep Graph Library (DGL)~\cite{wang2019dgl}, with default hyperparameters wherever required and keep them constant throughout. Analogous to a simple Neural Network (NN) with a hidden layer, we build a Graph Neural Network (GNN) model with 2 same convolution layers (hidden and output) for each baseline with a ReLU activation function in between. The hidden and final embedding dimensions are kept constant for each baseline model and that of PersonaSAGE for a fair comparison.
Though ours is a multi embedding approach, we keep the final embedding size equal to that obtained from baselines by adjusting the output dimension ($D$) and the number of personas ($K$) such that their concatenation yields a match.

\subsubsection{Training and Evaluation.} \label{sec:training-and-evaluation}
Given a graph dataset, we hold out $15\%$ of data for testing model performance while another $15\%$ to select the best model learned during training as part of the validation set. All models are trained for $100$ epochs using Adam Optimizer~\cite{kingma2014adam} with a learning rate of $0.01$. The loss function used for optimization is cross-entropy between predictions and ground truth values. All experiments are seeded for reproducibility and run with 5 different seeds resulting in different parameter initializations of weights along with a different data split. The mean and standard deviation of best scores are reported in tables. All other hyperparameters (hidden size, output size, etc.) are kept constant throughout. The embedding size for the first GNN layer is set to $128$. For Cora, the embedding size of second layer $D$ is set to $70$ for all baseline models while for PersonaSAGE it is $10$, since we assume $K=7$ personas in the data and each persona embedding is of size $10$, the concatenated embedding size ($7x10 = 70$) is unchanged. Similarly, for Citeseer we assume $K=6$ and for PubMed $K=3$ with $D=10$. Results with `$-$' indicate lack of completion. Unless otherwise mentioned, the PersonaSAGE variants use the mean aggregator. Further, PersonaSAGE refers to the default approach that uses the mean aggregator with Ward clustering.

\begin{figure}[b!]
    \centering
    \vspace{-2mm}
    \includegraphics[width=1.0\linewidth]{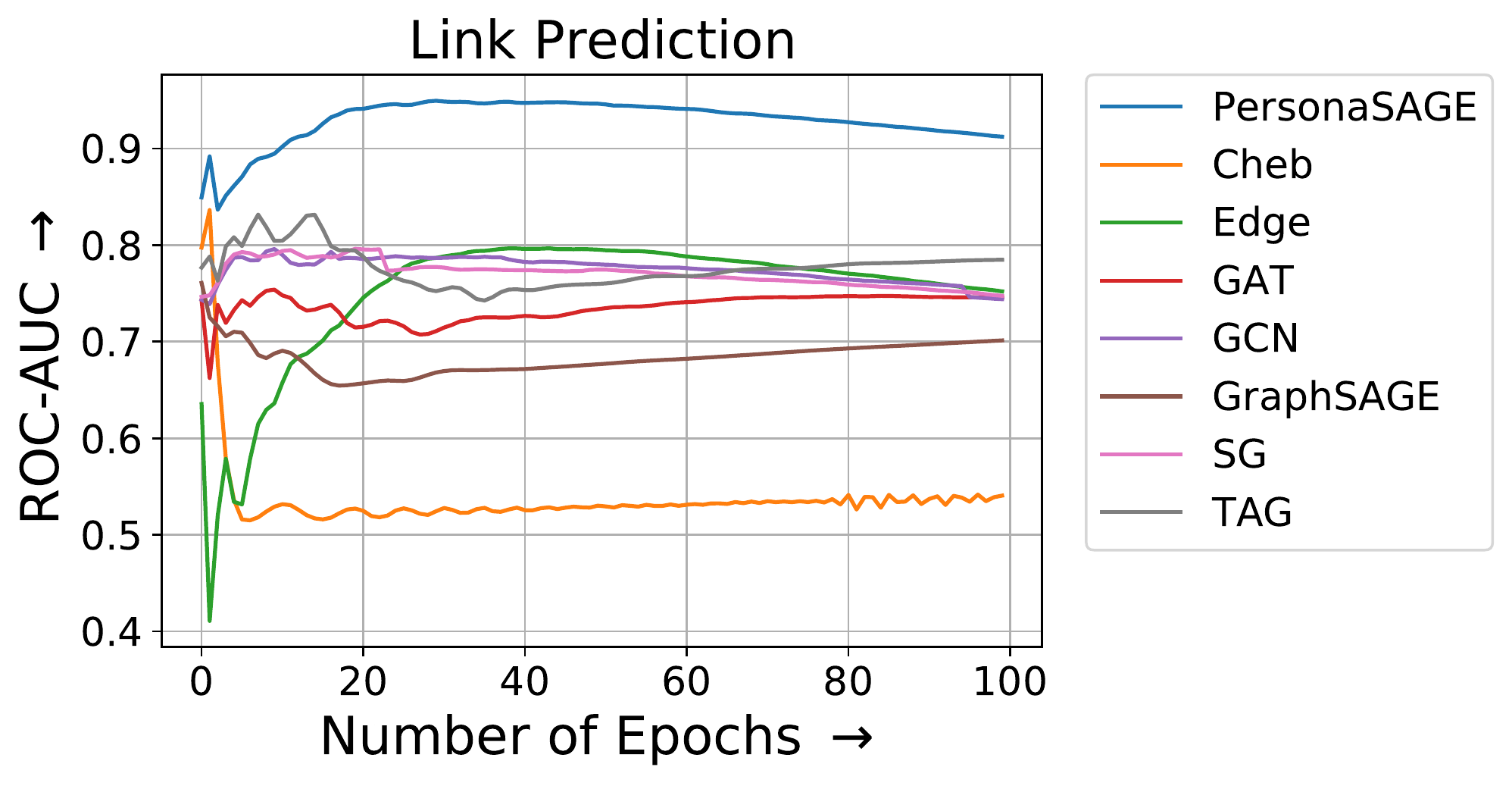}
    \vspace{-5mm}
    \caption{Comparing link prediction test performance (AUC) as a function of the number of epochs used for training on the PubMed dataset.
    Strikingly, PersonaSAGE achieves significantly better performance across all training epochs.
    }
    \label{fig:training-phase-pubmed-auc}
    \vspace{-2mm}
\end{figure}

\subsection{Link Prediction} \label{sec:link-prediction}
We first investigate the effectiveness of PersonaSAGE for link prediction.
In particular, the goal of the link prediction task is to accurately predict the likelihood of the formation of new edges in the graph. Given a pair of nodes, the likelihood can be computed in terms of similarity between the two nodes as the inner product of the two node embeddings followed by a Sigmoid activation. 
This task measures the capability of the embeddings to capture the structure and topology of the graph. 
The edge set is randomly split into train and test sets, thus serving as positive samples for each set. An equal size set of remaining unconnected edges is constructed and split in the same proportions as the train and test set. This randomly sampled set serves as negative samples.
This setup is adopted in many works, see~\cite{abu2017learning,deepGL,grover2016node2vec}.

In Table~\ref{tab:link-prediction-main}, we compare PersonaSAGE to a wide variety of other graph representation learning methods.
For link prediction, we investigate using both the raw input features that are typically used for node classification (due to their correlation with the class labels of the nodes) and we also study using random features since there is no guarantee that the raw input features would be useful for the link prediction task. 
Notably, we observe that PersonaSAGE outperforms the other methods across all graph datasets and across both feature settings including the setting where raw input features are used and another where we instead leverage random features (Table~\ref{tab:link-prediction-main}).
Interestingly, PersonaSAGE performs best on Citeseer and Cora when the input features are randomly initialized as opposed to using the standard input features that are typically used for node classification.
For instance, we achieve 0.907 on Citeseer when random features are used compared to only 0.872 when using the raw input features as shown in Table~\ref{tab:link-prediction-main}.
PersonaSAGE always outperforms the best baseline method with a gain of $6\%$, $15\%$, and $14\%$ using raw input features and $13\%$, $27\%$, and $12\%$ when random features are used for Citeseer, Cora, and Pubmed, respectively.
Nevertheless, in all cases, PersonaSAGE achieves significantly better predictive performance compared to the wide range of baseline methods. 
These results indicate the advantage and utility of PersonaSAGE over the other baselines for this prediction task.

\begin{table}[t!]
    \centering
    \vspace{-3mm}
    \caption{
    Node Classification Results.
    }
    \label{tab:node-classification-main}
    \vspace{-4mm}
    \begin{tabular}{l c c c}
    \toprule
    \textbf{\textit{Model}}& \textbf{Citeseer} & \textbf{Cora} & \textbf{PubMed} \\
    \midrule
    \textit{Cheb}             & 0.707 $\pm$ 0.04                           & 0.763 $\pm$ 0.04                     & 0.876 $\pm$ 0.0                        \\
	\textit{Edge}             & 0.629 $\pm$ 0.08                         & 0.719 $\pm$ 0.02                      & 0.862 $\pm$ 0.0                        \\
	\textit{GAT}              & 0.665 $\pm$ 0.03                          & 0.781 $\pm$ 0.02                       & 0.830 $\pm$ 0.0                         \\
	\textit{GCN}              & 0.721 $\pm$ 0.04                          & 0.851 $\pm$ 0.02             & 0.843 $\pm$ 0.01               \\
	\textit{GraphSAGE}   & 0.714 $\pm$ 0.03                          & 0.781 $\pm$ 0.04                     & 0.866 $\pm$ 0.01                       \\
	\textit{SG}               & 0.709 $\pm$ 0.04                         & 0.853 $\pm$ 0.02                    & 0.842 $\pm$ 0.01                       \\
	\textit{TAG}              & 0.616 $\pm$ 0.10                           & 0.833 $\pm$ 0.04                     & 0.872 $\pm$ 0.0                       \\
	\midrule
    \textit{PersonaSAGE} 
	& \textbf{0.722 $\pm$ 0.02}                        & \textbf{0.859 $\pm$ 0.01}                      & \textbf{0.877 $\pm$ 0.01}            \\
	\bottomrule
    \end{tabular}
    \vspace{-3mm}
\end{table}

In Figure~\ref{fig:training-phase-pubmed-auc}, we investigate the performance of the different methods as the number of training epochs varies.
Strikingly, PersonaSAGE achieves significantly better performance across all number of training epochs.
This holds true for training with very few epochs as observed in Figure~\ref{fig:training-phase-pubmed-auc}.
Furthermore, the best performance of PersonaSAGE is achieved when using only 20 training epochs, and slightly decreases with additional epochs.
In contrast, many of the other methods achieve their best performance using a far larger number of epochs compared to PersonaSAGE.
For instance, GraphSAGE achieves an AUC of 0.70 when using 100 training epochs whereas PersonaSAGE achieves a significantly better AUC of 0.90 using only 10 epochs.
This is a 10x difference in the number of epochs, while significantly outperforming GraphSAGE in terms of AUC (0.90 compared to 0.70).

\begin{table}[b!]
    \centering
    \vspace{-1mm}
    \caption{Ablation study results comparing different PersonaSAGE variants for node classification.
    }
    \label{tab:node-classification-ablation-study}
    \vspace{-4mm}
    \begin{tabular}{@{} l c c c@{}@{}@{}}
    \toprule
    \textbf{\textit{Model}}& \textbf{Citeseer} & \textbf{Cora} & \textbf{PubMed} \\
    \midrule
    \textit{PersonaSAGE-KM (K=1)}
	& 0.663 $\pm$ 0.07               & 0.837 $\pm$ 0.04                     & 0.872 $\pm$ 0.01                      \\
	\textit{PersonaSAGE-KM-max}      
	& 0.720 $\pm$ 0.02                           & 0.858 $\pm$ 0.01                      & \textbf{0.878 $\pm$ 0.0}                 \\
	\textit{PersonaSAGE-KM-sum}   
	& 0.699 $\pm$ 0.03                           & 0.850 $\pm$ 0.02                        & 0.728 $\pm$ 0.17                        \\
	\textit{PersonaSAGE-KM} 
	& 0.716 $\pm$ 0.03                           & 0.838 $\pm$ 0.02                      & 0.874 $\pm$ 0.01                        \\
    \textit{PersonaSAGE-Birch} 
	& 0.722 $\pm$ 0.02                          & \textbf{0.861 $\pm$ 0.01}              & 0.875 $\pm$ 0.01                       \\
    \textit{PersonaSAGE-Spec}
	& \textbf{0.729 $\pm$ 0.02}                 & 0.854 $\pm$ 0.02                      & 
	-
	           \\
	\textit{PersonaSAGE} 
	& 0.722 $\pm$ 0.02                         & 0.859 $\pm$ 0.01                      & 0.877 $\pm$ 0.01             \\
	\bottomrule
    \end{tabular}
    \vspace{-2mm}
\end{table}

In Table~\ref{tab:link-prediction-ablation-study}, we investigate a few different variants from the proposed PersonaSAGE framework. 
Notably, we observe that these different variants often outperform the previous variant in Table~\ref{tab:link-prediction-main}.
For instance, the PersonaSAGE-KM-max variant that uses K-Means (KM) with the max aggregator performs the best on PubMed.
Furthermore, while there is not a single PersonaSAGE variant that always performs best across all graphs and input features, the PersonaSAGE variant that uses k-means with the mean relational aggregator outperforms the other variants most consistently.
In particular, this variant always performs best when using the actual input features compared to using random input features (which models the case where such features may not exist for the input graph).
As an aside, we also compare a PersonaSAGE variant that leverages a single embedding per node called PersonaSAGE-KM (K=1), which as shown in Table~\ref{tab:link-prediction-ablation-study} often performs the worst compared to the other PersonaSAGE variants, indicating the effectiveness of learning multiple embeddings per user.
We perform additional ablation study experiments in Section~\ref{sec:ablation-study}.

\subsection{Node Classification}\label{sec:exp-node-classification}
Now we investigate using PersonaSAGE for the node classification task.
In Table~\ref{tab:node-classification-main}, we observe that PersonaSAGE always outperforms the other methods across all benchmark datasets.
For node classification, we use the raw input features since they are known to be correlated with the class labels we are predicting for the nodes.
We also investigate a number of PersonaSAGE variants from the proposed framework for node classification.
Results are provided in Table~\ref{tab:node-classification-ablation-study}.
Notably, we observe that different PersonaSAGE variants perform best for the different graphs.
In particular, PersonaSAGE-Spec performs best for Citeseer, PersonaSAGE-Birch performs best for Cora, and PersonaSAGE-KM-max performs best for PubMed.
Most importantly, in all cases, we find that these results are even better than the PersonaSAGE results in Table~\ref{tab:node-classification-main}.
Hence, these variants achieve even better performance compared to the other methods.

\begin{figure}[t!]
    \centering
    \vspace{-6mm}
    \includegraphics[width=1.0\linewidth]{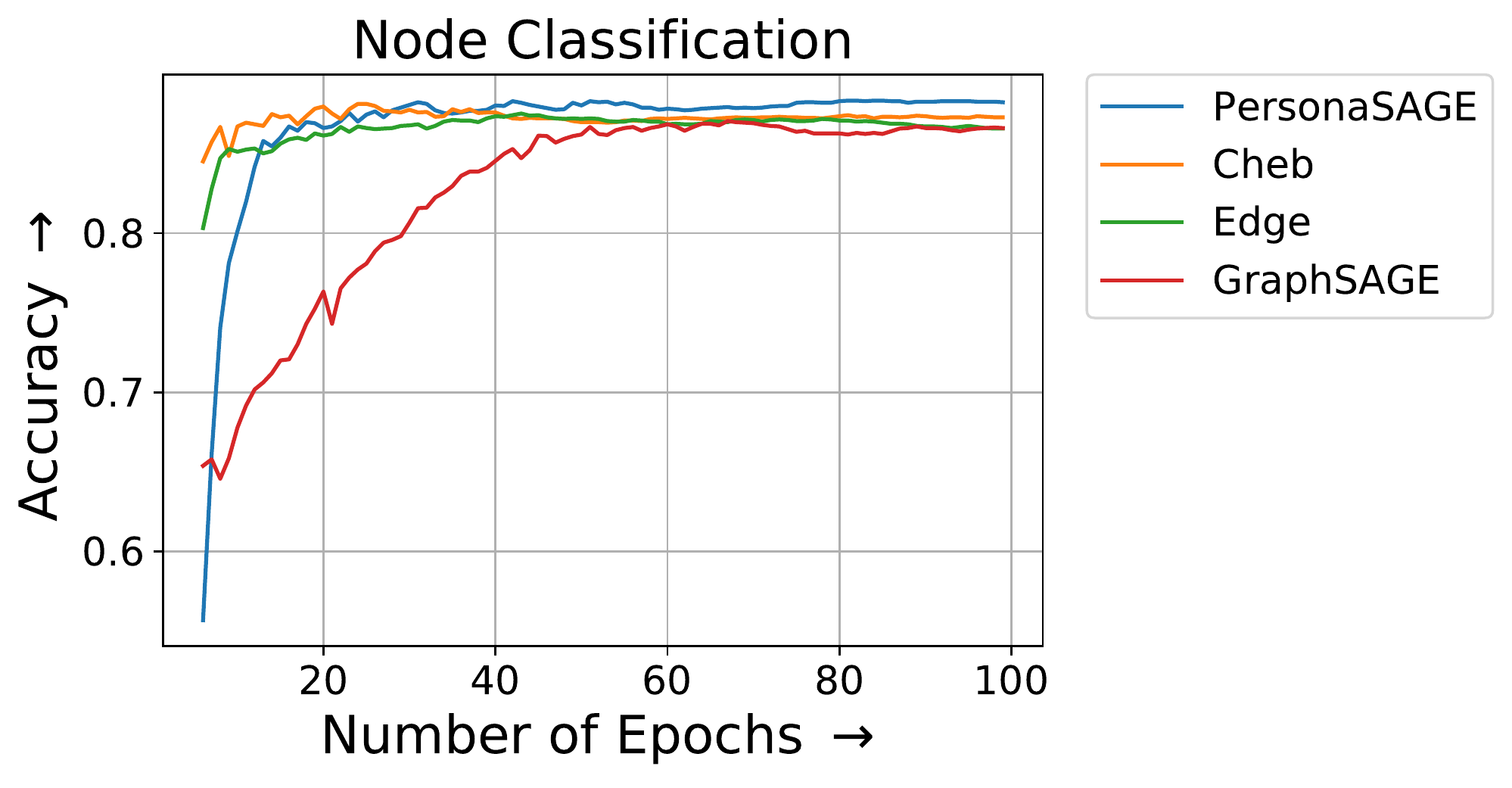}
    \vspace{-7mm}
    \caption{
    Comparing the node classification test accuracy as a function of the number of epochs on the PubMed dataset.
    Notably, PersonaSAGE outperforms the other methods even when a relatively small number of epochs is used for training.
    }
    \label{fig:node-classification-training}
    \vspace{-3mm}
\end{figure}

To investigate the effectiveness of PersonaSAGE as the number of epochs increases, we compare the node classification accuracy as the number of epochs increases for PersonaSAGE along with a variety of state-of-the-art methods.
From Figure~\ref{fig:node-classification-training}, we observe that PersonaSAGE achieves the best performance compared to the other methods, even when a modest number of epochs is used.
In particular, PersonaSAGE always outperforms the other methods when using 40 epochs or more as shown in Figure~\ref{fig:node-classification-training}.

\subsection{Ablation Study}\label{sec:ablation-study}
\begin{figure}[t!]
    \centering
    \vspace{-4mm}
    \begin{subfigure}[b]{\linewidth}
        \centering
        \includegraphics[width=.8\linewidth]{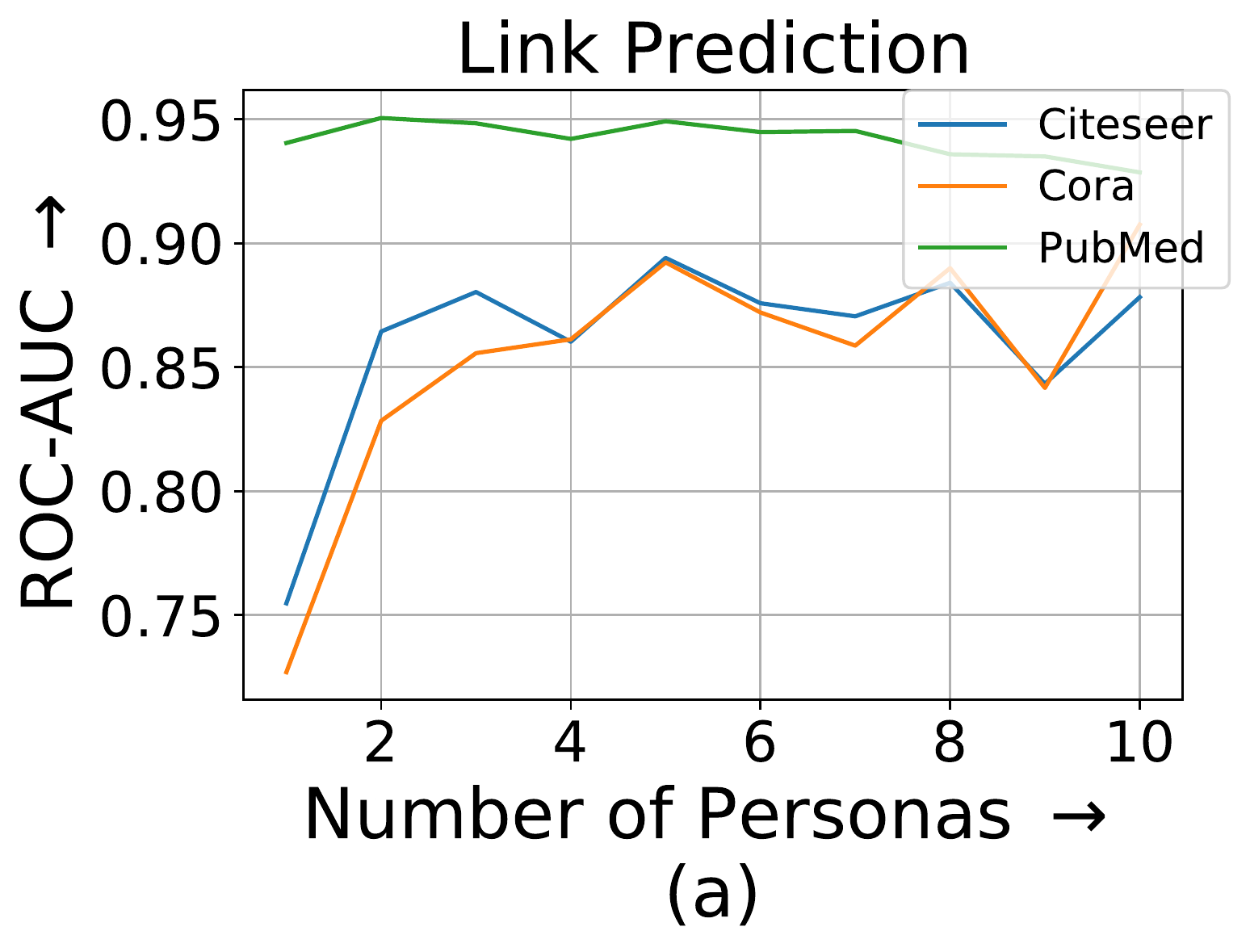}
    \end{subfigure}%
    \\[\baselineskip]
    \vspace{-2mm}
    \begin{subfigure}[b]{\linewidth}
        \centering
        \includegraphics[width=0.8\linewidth]{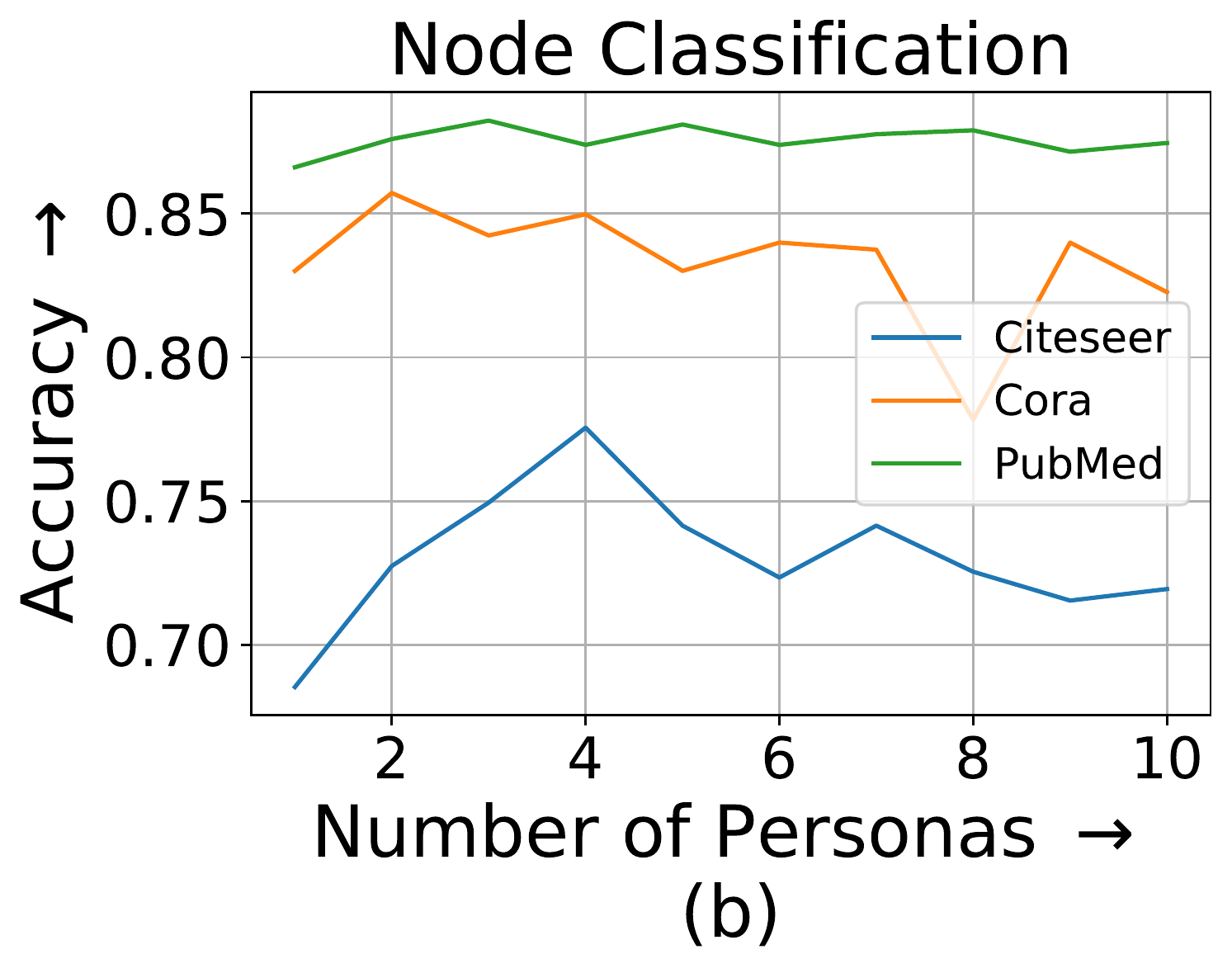}
    \end{subfigure}
    \vspace{-6mm}
    \caption{
    Varying the maximum number of persona embeddings $K$ per node in PersonaSAGE for link prediction and node classification.
    }
    \label{fig:k-variation}
    \vspace{-3mm}
\end{figure}

We investigate the impact of the maximum number of persona embeddings per node $K$ by varying this hyperparameter from $K \in \{1,2,\ldots,10\}$.
For this experiment, we use the PersonaSAGE variant that leverages K-Means and mean aggregator (PersonaSAGE-KM).
We ensure the final embedding size remains constant by reducing the size of the persona embeddings accordingly.
For instance, suppose the final embedding size is 50 and the maximum number of embeddings per node is $K=10$, then the size of each of the $K=10$ embeddings per node is set to be of size $D=5$.
Results are reported in Figure~\ref{fig:k-variation} for both link prediction and node classification.
Overall, we observe that for link prediction, the performance generally increases for Cora and Citeseer while remaining approximately the same for PubMed.
In contrast, node classification performance of Citeseer is best when $K=4$, then decreases as $K$ increases further, whereas the best performance of PersonaSAGE on Cora is $K=2$ and remains approximately the same for Citeseer. 
These results indicate the utility of PersonaSAGE and its ability to learn multiple embeddings per node in the graph.

\section{Case Study}
\label{sec:case-study}
In this section, we investigate using PersonaSAGE to recommend different types of entities from usage log data of a data management system. Such user interaction logs contain the data queries executed, datasets used, along with the attributes selected in those datasets. By leveraging such usage logs of user activities, we can learn a model for the personalized recommendation of such entities. The goal is to leverage user interactions (from the usage logs) with the platform for building a personalized recommendation engine for users, datasets, queries, etc.

\subsection{Data and Graph Construction}
The usage log data (from a data management system) consists of a list of queries issued by users interacting with a data management platform. Each query consists of a list of attributes (columns) referenced from their respective datasets. We parse the query to yield its referenced attributes and datasets using SQL Parser (Python Library) and map it with the corresponding user along with the issued query. We consider a heterogeneous graph with 4 types of nodes including users, queries, datasets, and attributes. We form an (undirected) edge between, for instance, a user and query node if the user has initiated that query. Similarly, we form edges between users and datasets, users and attributes based on interactions.

\subsection{Experimental Setup}
We investigate the persona embeddings for this large and sparse heterogeneous graph in the context of link prediction. The task designed is exactly same as stated in Section~\ref{sec:link-prediction} just that the whole graph is treated as homogeneous and the negative samples are drawn uniformly across the graph. This set would contain unconnected node pairs from all combinations such as user-user, user-query, query-dataset, and so on. Further, for conducting the experiment, we use a two-layer GNN architecture with the same hyperparameters as described in Section~\ref{sec:training-and-evaluation} unless explicitly stated. Since we do not have input features, we randomly initialize all of them with embedding size $100$. The hidden size is kept $128$ and output dimension $D$ is varied over $\{ 20, 30, 50, 70, 100\}$.

\begin{figure}[t!]
    \centering
    \vspace{-2mm}
    \includegraphics[width=1.0\linewidth]{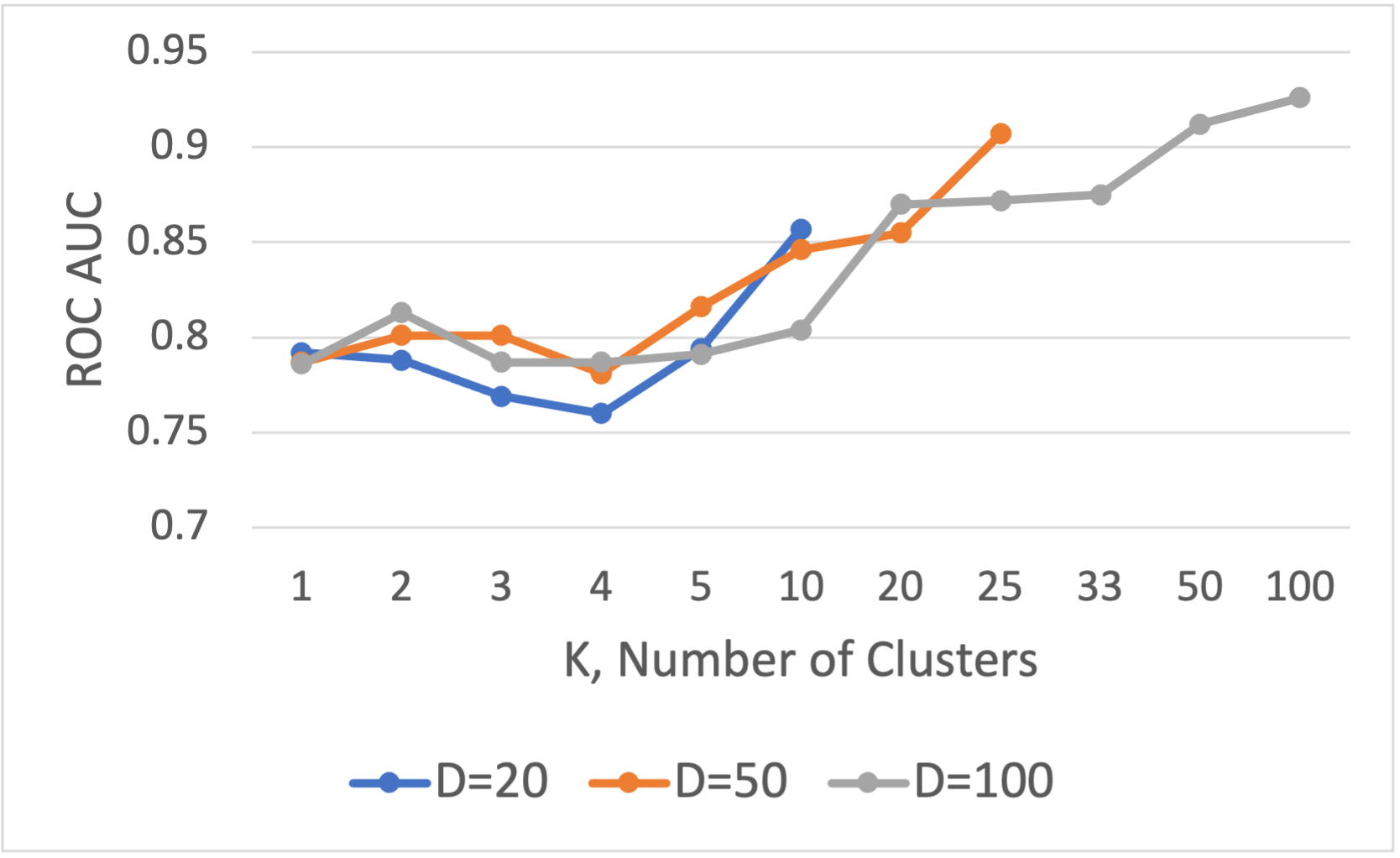}
    \vspace{-4mm}
    \caption{Personalized recommendation performance (ROC AUC) as a function of the max number of persona embeddings per node (number of clusters) $K$ 
    for the usage log data (from a data management system).
    }
    \label{fig:vary-clusters}
    \vspace{-2mm}
\end{figure}

\begin{table}[h!]
    \centering
    \vspace{-1mm}
    \caption{Results for Personalized Query, Attribute, and Dataset Recommendation (AUC).
    }
    \vspace{-4mm}
    \label{table:case-study-rec}
    \begin{tabular}{l c c c}
    \toprule
    & \textit{\textbf{Query}} & \textit{\textbf{Attribute}} & \textit{\textbf{Dataset}} \\
    \midrule
    Cheb                     & 0.500            & 0.500                &  0.500             \\
    Edge                                 & 0.668                   & 0.684                       & 0.548                     \\
    GAT                                  & 0.593                   & 0.495                       & 0.689                     \\
    GCN                                  & 0.700                     & 0.730                        & 0.635                     \\
    GraphSAGE                            & 0.723                   & 0.716                       & 0.660                      \\
    SG                                   & 0.683                   & 0.655                       & 0.604                     \\
    TAG                                  & 0.722                   & 0.707                       & 0.579                     \\
    \midrule
    \textbf{PersonaSAGE} & 
    \textbf{0.862}          & \textbf{0.845}              & \textbf{0.839}    \\
    \midrule
    \end{tabular}
    \vspace{-2mm}
\end{table}

\subsection{Results}
In this section, we investigate using PersonaSAGE to recommend any arbitrary type of entity to a user in a personalized fashion.
In particular, results for three different personalized recommendation tasks including \textsc{query}, \textsc{attribute}, and \textsc{dataset recommendation} are provided in Table~\ref{table:case-study-rec}.
Overall, we observe that PersonaSAGE outperforms the other methods across all three personalized recommendation tasks as shown in Table~\ref{table:case-study-rec}.
The improvement compared to the next best method across the three different recommendation tasks is significant.
Notably, compared to the best performing baseline method, PersonaSAGE achieves a gain of $19.2\%$, $15.7\%$, and $21.7\%$ in AUC  for query, attribute, and dataset recommendation tasks, respectively. Furthermore, while PersonaSAGE always outperforms the other methods across all three recommendation tasks, there is not a single baseline method that always performs best across all three recommendation tasks.
For instance, GraphSAGE is the best performing baseline method for query recommendation, whereas GAT outperforms the other baseline methods for the dataset recommendation task.
In contrast, PersonaSAGE performs best independent of the personalized recommendation task.

To further understand the approach, we now investigate a slightly different experimental setup where the held out set for a specific user can contain a mix of different entity types, including datasets, attributes, and queries.
Results are provided in Table~\ref{table:case-study-results}. 
Notably, we observe that PersonaSAGE outperforms the other methods across all embedding sizes $D \in \{20,30,50,70,100\}$ as shown in Table~\ref{table:case-study-results}.
Furthermore, PersonaSAGE achieved a mean gain of $12.37\%$ in AUC over the best performing baseline when $D=100$ is used.
In Figure~\ref{fig:vary-clusters}, we report performance as we vary the maximum number of persona embeddings per node for $D\in \{20,50,100\}$.
Notably, these results indicate that PersonaSAGE is well-suited to predict any arbitrary link type, as we do not restrict the held-out links to a specific type in this experiment.
These results demonstrate the overall effectiveness of our approach for prediction in such complex heterogeneous graphs.

\begin{table}[h!]
    \centering
    \vspace{-1mm}
    \caption{
    Results for recommendation of any arbitrary link-type (AUC). 
    These results include recommendation of a variety of different link-types such as user-query links, user-dataset links, and user-attribute links. 
    }
    \vspace{-3mm}
    \label{table:case-study-results}
    \begin{tabular}{l c cc cc}
    \toprule
    & \textbf{$D=20$}    & \textbf{$D=30$} & \textbf{$D=50$} & \textbf{$D=70$} & \textbf{$D=100$}    \\
    \midrule
    \textit{GraphSAGE}                    & 0.770        & 0.770        & 0.778       & 0.755       & 0.778        \\
    \textit{GCN}                            & 0.828       & 0.803       & 0.72        & 0.837       & 0.822        \\
    \textit{TAG}                            & 0.787       & 0.848       & 0.854       & 0.809       & 0.824        \\
    \textit{GAT}                            & 0.725       & 0.75        & 0.687       & 0.731       & 0.729        \\
    \midrule
    \textit{PersonaSAGE}        & \textbf{0.857}       & \textbf{0.861}       & \textbf{0.907}       & \textbf{0.909}       & \textbf{0.926}    \\
    \midrule
    \end{tabular}
    \vspace{-3mm}
\end{table}

\section{Related Work} \label{sec:related-work}
Recent advances in approaches for network embeddings have received a lot of attention due to their effectiveness in capturing both local and global contexts of the networks. These approaches for learning network representations can be grouped based on various criteria. 
Some of the graph convolution methods include
GCN~\cite{kipf2016semi} and Chebnet~\cite{defferrard2016convolutional}, which present a model based on spectral convolution graph operations, extending convolutions from Euclidean grids to graphs. The forward propagation is computed by stacking layers of the product of a normalized graph Laplacian, the data, and the model parameters, which is the result of approximating Chebyshev polynomials. These papers utilize techniques to optimize computation, and report results across distinct datasets. In \cite{defferrard2016convolutional} results are presented for the image classification dataset, and text categorization dataset. \cite{kipf2016semi} showcases results for a 2-layer GCN node classifier on citation network datasets. 

Apart from spectral-based approaches~\cite{henaff2015deep}, there are various spatial-based convolution approaches~\cite{hamilton2017inductive, monti2017geometric} for effectively leveraging the spatial contexts. For instance, Graph Attention Networks (GAT)~\cite{velivckovic2017graph} implements multi-head attention on the local neighborhood of a node and demonstrates results on citation networks, and a protein-protein interaction dataset.
Edgeconv \cite{wang2019dynamic} builds models for the tasks of classification, part segmentation, and semantic segmentation through edge convolutions on point cloud data.
SGN \cite{wu2019simplifying} simplifies the architecture of GCN by removing non-linearities, and presents results on citation and social networks.
A survey of several graph neural network architectures is presented in \cite{wu-gnn-survey}.

Another line of works learn node representations using simulated random walks on the graph. Works by~\cite{perozzi2014deepwalk, grover2016node2vec, tang2015line, role2vec}, which compute (single) node embeddings based on random walks are transductive approaches in nature. They require a retraining procedure every time a new node is encountered. There have been some inspiring recent advances in the direction of capturing the polysemous behavior of nodes in a graph. PolyDeepwalk~\cite{liu2019single} propose a multi-embedding approach to model multiple \textit{facets} of nodes and highlight the effectiveness of a multi-embedding representation than a single vector. Asp2vec~\cite{park2020unsupervised} propose an unsupervised end-to-end pipeline to compute multiple \textit{aspects} of nodes based on their local context. However, these approaches follow the same drawback of using random walks which render them less effective in scenarios for growing networks.
PinnerSage~\cite{pal2020pinnersage} is a bipartite clustering approach that assigns users multiple embeddings.
However, the embeddings assigned to the users are not unique, that is, two or more users can share the exact same embedding vector, which makes the multiple embeddings not as useful. 
Other work introduced polysemy embeddings~\cite{epasto2019single} but simplify the problem by learning embeddings for each facet (or graph). This is essentially equivalent to heterogeneous embedding methods that learn embedding vectors for each context/mode.

All these works inspire our thinking for developing a flexible, general, and yet performant framework for capturing polysemous nature (persona) in a large-scale network. 
We note a differentiation of our work from these multi-embedding approaches is that our approach learns a \emph{set} of embeddings where the \textit{number} of (persona) embeddings may be different for different nodes which are learned automatically from local and global structures.

\section{Conclusion} \label{sec:conc}
We generally come across scenarios where a single entity performs in a polysemous way, such as an individual's behavior in the context of a sports player, father, etc. We represent this nature as multiple personas of the same entity but in different contexts. In this work, we proposed a novel approach called PersonaSAGE that learns multiple persona embeddings per node along with their persona weights. Notably, the set of embeddings learned for every node may differ depending on the structural context around a given node. We demonstrated the effectiveness of PersonaSAGE for a wide variety of application tasks including node classification and link prediction. Finally, we also conducted a case study where we investigated using PersonaSAGE for recommending queries, attributes, and datasets to users. In all cases, PersonaSAGE significantly outperformed the other methods across all graphs and application tasks.

\balance
\bibliographystyle{ACM-Reference-Format}
\bibliography{paper}

\appendix
\section*{Appendix}

\section{Experimental Setup Details} \label{sec:exp-setup-details}

\subsection{Data}
The datasets used in the experiments are described below:

\begin{itemize}
% reference: http://zhang18f.myweb.cs.uwindsor.ca/datasets/#h4.5_pubmed-a-href-.-pubmed.tar.gz-download-a-
\item Each node (a publication) in the Citeseer dataset is labeled as one of the six domains: Agents, Artificial Intelligence, Databases, Information Retrieval, Machine Learning, and Human Computer Interaction. 

\item The Cora dataset consists of publications (nodes) belonging to one of the following classes: Case Based, Genetic Algorithms, Neural Networks, Probabilistic Methods, Reinforcement Learning, Rule Learning, and Theory.

\item The PubMed dataset comprises of publications (nodes) classified as one of the $3$ categories of Diabetes (Experimental, Type 1 and Type 2).

\end{itemize}

\subsection{Baselines}
The baselines used in experiments are described below:
\begin{itemize}
    \item ChebNet (Cheb)~\cite{defferrard2016convolutional}: The Chebyshev spectral filter, derived from the Spectral CNN, uses an efficient pooling strategy to capture the features in a localized region.
    \item Graph Convolution Network (GCN)~\cite{kipf2016semi}: This kind of convolution operation introduced a first-order approximation of the ChebNet in the space of spectral filters. It can be considered as combining information from neighbors with a self-loop on every node.
    \item GraphSAGE~\cite{hamilton2017inductive}: The authors proposed an inductive approach to aggregate neighborhood information with the node's current information and apply linear transformations followed by non-linearity (\eg, Sigmoid).
    \item Graph Attention Networks (GAT)~\cite{velivckovic2017graph}: Unlike GraphSAGE, this approach weighs the neighborhood information by learning the relative edge weights using an attention network.
    \item Topology Adaptive Graph convolutional networks (TAG)~\cite{du2017topology}: The authors propose a topology-aware spectral filter for graph convolution operation by extending the neighborhood definition used in GCN to capture higher-order neighborhood information.
    \item EdgeConv (Edge)~\cite{wang2019dynamic}: Inspired from PointNet~\cite{qi2017pointnet}, the authors propose a differential and pluggable convolution operator module designed to capture the topological and geometric features of point clouds.
    \item Simple Graph convolution (SGN)~\cite{wu2019simplifying}: The authors propose a rather simplifying computation for graph convolution operation by reducing the non-linearities in order to scale to large graphs and without drastically reducing the performance.
\end{itemize}

\section{Properties of PersonaSAGE} 
In this subsection, we note some salient aspects of the PersonaSAGE algorithm by remarking on the variation in the number of persona embeddings per node.

{\textbf{Variable number of persona embeddings}}: 
For an arbitrary graph, the number of embeddings per node will depend on the clustering algorithm and topology of the graph. For instance, consider a graph with $n$ nodes and $k=n$ clusters, where start and end nodes have degree $1$ and interior nodes have degree $2$ and are connected to the previous and next nodes.
In this case, if the number of GNN layers is less than $\log_3 n$, then there will be a variable number of persona embeddings for each node, which can be obtained by analyzing the aggregation of the membership vectors across the layers.

{\textbf{Weisfeiler-Lehman test}}:
The GraphSAGE algorithm is related to the Weisfeiler-Lehman test.
This is a graph isomorphism test, which maintains node labels or hashes by aggregating labels of neighboring nodes, and across iterations.  
The $1$-WL test, which is also called the vertex color refinement algorithm, aggregates information for a node across its node neighbors. The $k$-dimensional Weisfeiler-Lehman test defines the node neighborhood to be $n$ many $k$-tuples of nodes, whereas the $k$-FWL version considers $k$ number of $k$-tuples.
It is known that for each $k \geq 2$ there is a pair of non-isomorphic graphs distinguishable by $(k + 1)$-WL but not by $k$-WL. Higher-order variants of the WL-test form the basis of recent work on graph neural networks \cite{gnn-wl-maron, chen-gnn-neurips-19, damke-gnn, morris-wl-sparse-neurips-20}. Though there are similarities with the WL-test, we do not consider the graph isomorphism problem in this paper further. 
In comparison to the prior art, $k$-dimensional cluster membership provides a distinct definition of a node's neighborhood, which determines the aggregation logic in the PersonaSAGE algorithm.

\end{document}